\documentclass[preprint,12pt]{elsarticle}

\usepackage{amsmath}
\usepackage{subcaption}
\usepackage{url}
\usepackage{cleveref}

\DeclareMathOperator*{\argmax}{argmax}
\usepackage{xspace}

\newcommand{\ie}{i.e.,\xspace}

\newcommand{\system}{TTR\xspace}
\newcommand{\densecap}{DenseCap\xspace}

\journal{Robotics \& Autonomous Systems}

\begin{document}

\begin{frontmatter}



\title{Talk-to-Resolve: Combining scene understanding and spatial dialogue to resolve granular task ambiguity for a collocated robot}


\author{Pradip Pramanick, Chayan Sarkar, Snehasis Banerjee, \\and Brojeshwar Bhowmick}

\address{Robotics and Autonomous Systems, TCS Research, India}


\begin{abstract}
The utility of collocating robots largely depends on the easy and intuitive interaction mechanism with the human. If a robot accepts task instruction in natural language, first, it has to understand the user's intention by decoding the instruction. However, while executing the task, the robot may face unforeseeable circumstances due to the variations in the observed scene and therefore requires further user intervention. In this article, we present a system called Talk-to-Resolve (TTR) that enables a robot to initiate a coherent dialogue exchange with the instructor by observing the scene visually to resolve the impasse. Through dialogue, it either finds a cue to move forward in the original plan, an acceptable alternative to the original plan, or affirmation to abort the task altogether. To realize the possible stalemate, we utilize the dense captions of the observed scene and the given instruction jointly to compute the robot's next action. We evaluate our system based on a data set of initial instruction and situational scene pairs. Our system can identify the stalemate and resolve them with appropriate dialogue exchange with 82\% accuracy. Additionally, a user study reveals that the questions from our systems are more natural (4.02 on average on a scale of 1 to 5) as compared to a state-of-the-art (3.08 on average).

\end{abstract}


\end{frontmatter}



\section{Introduction}
The idea behind the collocated robot is to employ them in various activities where they can lend a helping hand and make our living/workspace simpler and more coherent~\cite{milliez2018buddy, pramanick2018defatigue}. Though the number of robots in our daily surroundings is increasing over the last decade, their usability remains restricted due to a lack of an intuitive interaction interface, especially for non-expert users. As natural language interaction increases the acceptability and usability of a robot, a large number of research efforts have focused on enabling natural human-robot interaction~\cite{liu2019review}. 

\begin{figure}
    \centering
    \includegraphics[width=0.85\linewidth]{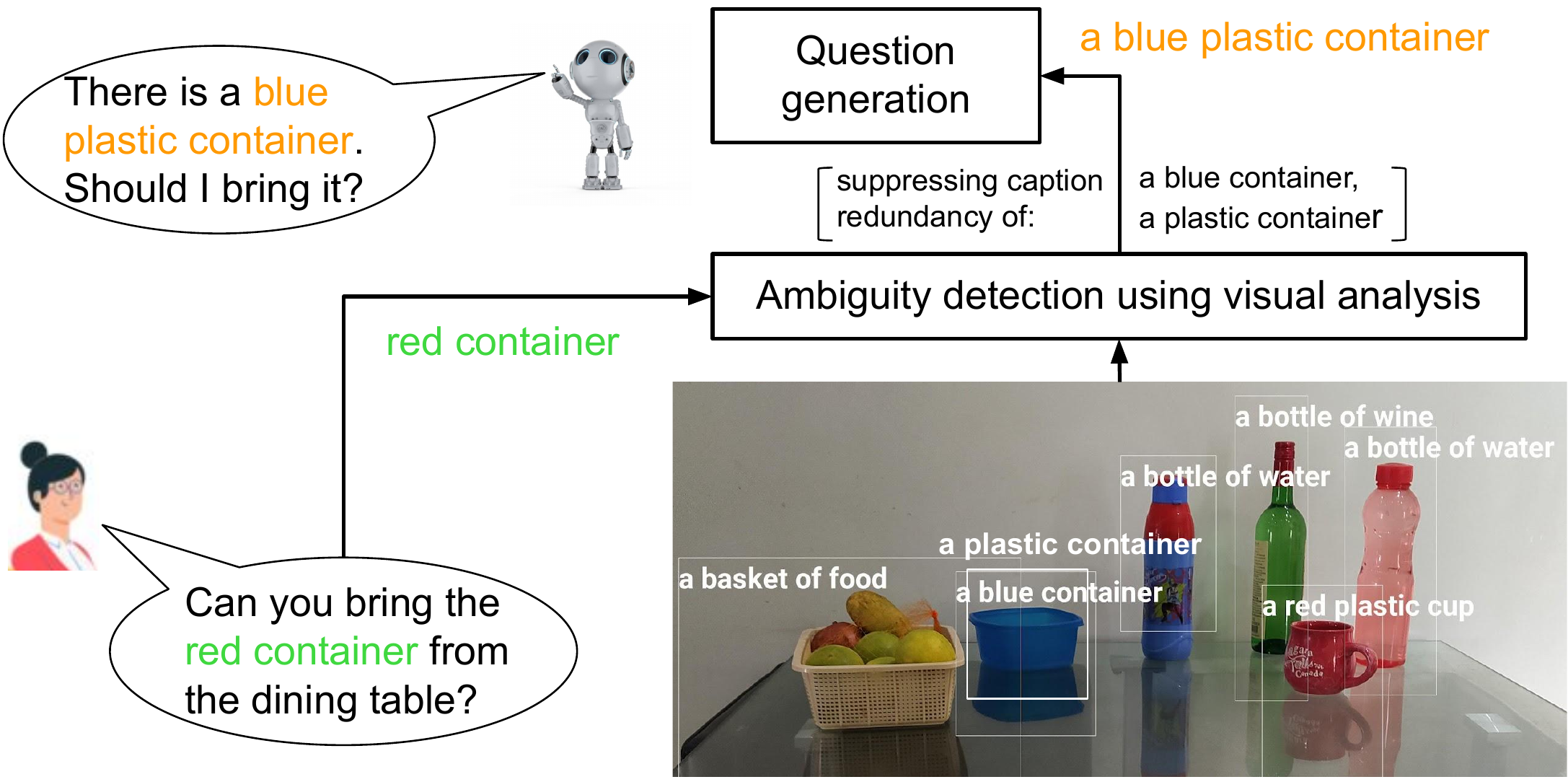}
    \caption{Representational image -- given the initial instruction, how TTR generates a suitable question after analyzing the scene when it faces task ambiguity.}
    \label{fig:representational}
\end{figure}

Figure~\ref{fig:representational} depicts a real-life scenario in a home environment where a fellow human being can ask a robot to perform certain tasks. In this case, she asks to bring the red container from the dining table. Assuming that the robot knows the environment, it first moves to the location of the dining table. The user instructs the robot based on the presumption that the red container is on the dining table. However, there is no guarantee that the mentioned object is at the desired location or is the only object at the location. So, first, the robot has to identify the exact and/or alternate entities in the scene to decide further course of action accordingly. In this case (Figure~\ref{fig:representational}), the robot can find only a blue plastic container at the location whereas the user has asked for a red one. As a result, the robot asks the user if the closest alternative is acceptable, i.e., should it bring the ``blue plastic'' container. In this work, we focus on enabling a robot with verbose ambiguity resolution capability. Given a task instruction in natural language and a scene from a robot's ego-view, the robot either generates an execution plan if doable in the current scenario or engages in a dialogue if there is any ambiguity in understanding what action is to be performed. As there can be different types of ambiguities/difficulties depending on the context, the first challenge is how to detect the nature of the problem the robot is facing. The subsequent question is how to convey the veracity of the problem to fellow human beings and seek guidance or direction.

In this work, we present Talk-to-Resolve (TTR), a multimodal natural interaction interface for robots for task ambiguity resolving. We utilize a state-of-the-art dense-caption generation system~\cite{johnson2016densecap} as the primary level of scene understanding. However, existing caption generation systems do not consider any instruction as a precursor to the scene analysis. As a result, captions are generated about every possible object that is available in the scene. Often multiple captions are generated about the same object, where each caption may provide a different description of the object. We perform caption merging and filtering operations to narrow down our search for detecting the desired object or location mentioned in the instruction. This step includes all the possible alternatives of the specified entities if there are multiple occurrences of the same object. After caption filtering and merging, we determine the level of ambiguity in the scene concerning the initial instruction. We have judiciously defined dialogue templates for each ambiguous state and generate a dialogue based on the predicted ambiguous state. 

To the best of our knowledge, this is the first work that generates contextual and informative questions during task execution to resolve both ambiguity and visual uncertainty by performing scene analysis. The main contributions of this work are three-fold.

\begin{itemize}
    \item Given a natural language description about an object and a scene understanding, we develop a novel method to identify the relevant object(s) while suppressing the redundant information efficiently.
    \item We meticulously design a set of ambiguity states and develop a dialogue system that resolves all possible types of ambiguity in scene understanding.
    \item Our dialogue system asks questions in a natural way that ensures that the user can understand the type of confusion the system is facing.
\end{itemize}
\section{Related work}
Executing natural language instruction given to a robot is a well-studied problem, particularly for object fetching and navigational instruction. However, the existing works in the literature mostly focus on instruction understanding for plan generation~\cite{pramanick2019enabling,dongcai17integrating} and assume that a generated plan can be executed without failure or further human intervention. Natural language instructions are prone to ambiguity and incompleteness that are often tackled using dialogue~\cite{thomason2019improving} and knowledge-based reasoning~\cite{chen2020enabling}. However, these systems only focus on the linguistic information provided by the human and do not take the uncertainty of the robot's perception into account when attempting to execute a plan. For example, in our earlier works, we have handled the natural language task instruction parsing to generate a high-level execution plan for the robot~\cite{pramanick2019enabling, pramanick2020decomplex}. These systems are supported by a dialogue engine that can raise a suitable query for the human user if the robot could not understand the task~\cite{pramanick2019your}. However, the robot would fail if the referenced object in the task cannot be identified uniquely (due to ambiguity) while executing the task. 

In practice, a robot may encounter unexpected situations during the execution of a plan, despite understanding the meaning of the instruction. To tackle this, a visual understanding of the environment, concerning the linguistic input, shows a promising direction. Recent works in vision and language navigation~\cite{mattersim} and object manipulation~\cite{misra2018mapping} can handle complex instructions using multi-modal information, but they still suffer from ambiguity and cascading errors due to misprediction. Although several works have specifically focused on the visual grounding of natural object descriptions~\cite{cohen2019grounding,magassouba2019understanding,sadhu2019zero,banerjee2021ontoscene}, they do not tackle ambiguity and incompleteness using dialogue. Moreover, the predominant approach of end-to-end training for visual grounding is difficult to use in a dialogue system, because the generation of a question pertaining to the instruction, requires a finer-grained understanding of the scene. Although visual question-answering systems can perform fine-grained scene analysis~\cite{johnson2017clevr, teney2018tips}, they are limited to answering questions, as opposed to generating a specific question to execute an instruction in a given scenario. Also, existing visual question generation models cannot be directly integrated into a robotic system, because they only generate natural questions about a given scene~\cite{patil2020visual,zhang2018goal}, irrespective of any given instruction. Moreover, such visual dialogue systems use the feedback from multiple questions to arrive at the conclusive question, which is undesirable in a human-robot dialogue for task execution, where the scope of asking multiple questions is limited.

\begin{table}[t]
    \centering
    \begin{tabular}{|c|p{2.1cm}|p{2.2cm}|p{3.5cm}|}\hline
        Method & Ambiguity detection in grounding & Granular ambiguity states & Query on \newline ambiguity \\ \hline
        FETCH-POMDP~\cite{whitney2017reducing} & implicit & binary \newline ambiguity & non-informative, fixed query \\ \hline
        Interactive Picking~\cite{hatori2018interactively} & explicit & binary \newline ambiguity & non-informative, fixed query \\ \hline
        INGRESS~\cite{shridhar18} & explicit & binary \newline ambiguity & informative, \newline but fixed query \\ \hline
        INVIGORATE~\cite{zhang2021invigorate} & explicit & binary \newline ambiguity & informative, \newline but fixed query \\ \hline
        \textbf{\system} (our system) & explicit & \textbf{multi-level ambiguity} & \textbf{informative, \newline contextual query} \\ \hline
    \end{tabular}
    \caption{Comparison of \system with respect to the state-of-the-art systems for ambiguity handling.}
    \label{tab:sota}
\end{table}

On the other hand, existing dialogue systems for robotic instruction understanding, mostly focus on eliciting missing information~\cite{thomason2019improving, pramanick2019your} and interactive task semantics learning~\cite{she2017interactive}, but do not tackle visual ambiguity and inconsistency. The most relevant works only focus on grounding and use dialogue in a very limited scope. Table~\ref{tab:sota} lists the closest works with respect to the problem that is tackled in this article. Whitney \textit{et al.}~\cite{whitney2017reducing} proposed a POMDP based object fetching task where pointing gestures are used to tackle ambiguities arising from open-ended instructions. They implicitly handle only binary type ambiguity, i.e., is there one object (non-ambiguous) or more object (ambiguous), and raise the same query if it is ambiguous. Hatori \textit{et al.} proposed a system for interactive picking that uses dialogue to resolve ambiguity in a picking task. However, the dialogue system tackles only binary ambiguity and it generates generic and open-ended questions such as ``which one?'', which often leads to further ambiguities. Shridhar \textit{et al.} proposed a similar system called INGRESS~\cite{shridhar18} where a referential expression generation technique is used to generate questions where binary answers are possible. However, their system only considers two dialogue states, i.e., the instruction is either ambiguous or completely understood (binary ambiguity). Recently, Zhang \textit{et al.} proposed an improved system called INVIGORATE~\cite{zhang2021invigorate} that combines multiple neural network systems using a POMDP model to tackle the uncertainty of each system jointly. However, their system is limited to binary ambiguity only; hence it restricts the usability of the system in a generic setup. In particular, the system would either lead to many rounds of question-answering with the human being or fail to resolve the ambiguity beyond the restrictive setup. In contrast, our system can resolve multiple variants of ambiguity. Also, it asks specific questions by conveying the robot's understanding of the scene that are easier to answer correctly (informative, contextual query).  

To the best of our knowledge, none of the existing works specifically focus on generating questions to complete a task by performing scene analysis, where the questions are asked to resolve both ambiguity and visual uncertainty at a granular level. Moreover, our system is more accurate in predicting the ambiguity states and generating more natural questions.

\section{System Overview}
\begin{figure}
    \centering
    \includegraphics[width=\linewidth]{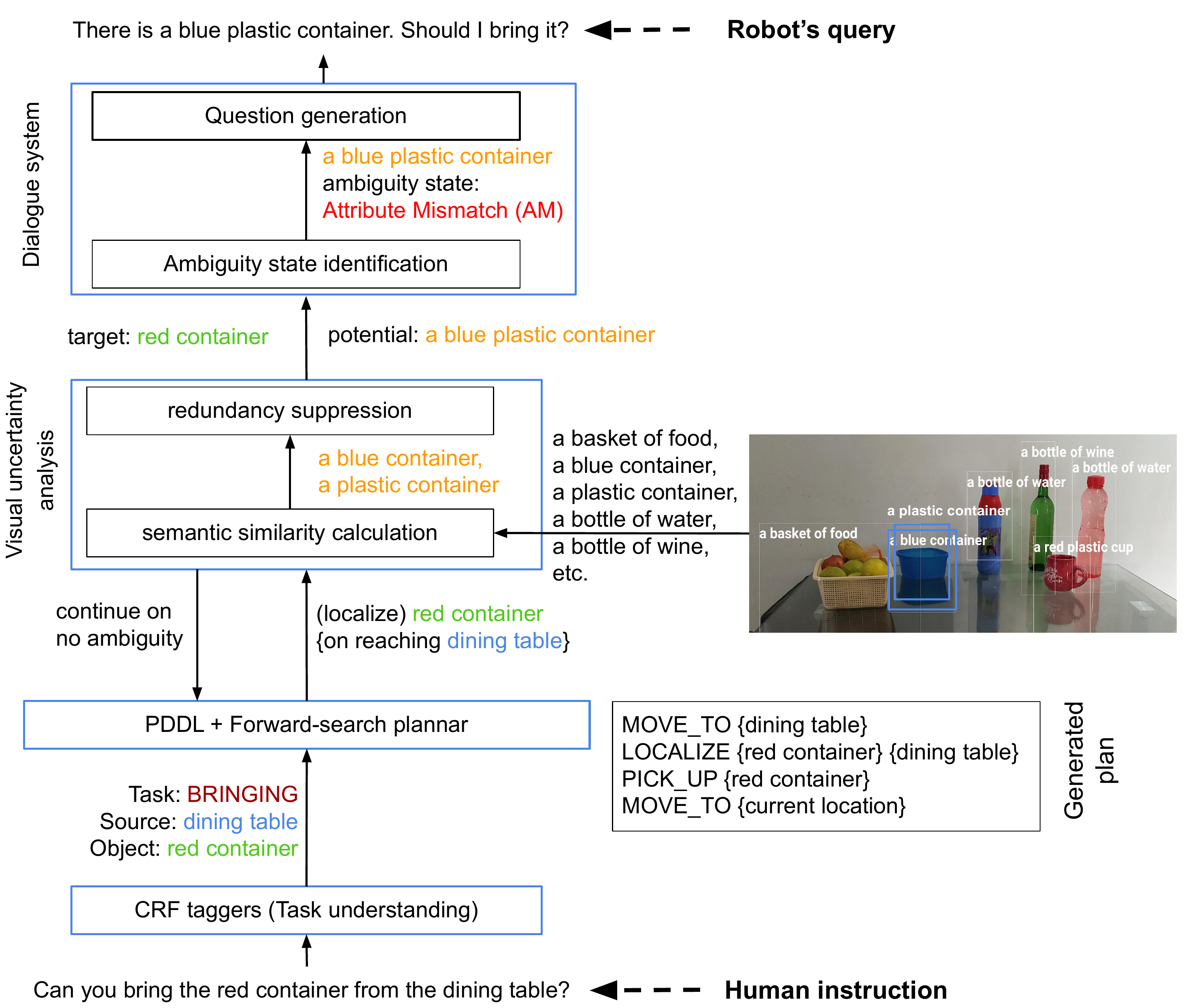}
    \caption{Overview of TTR showing how the major building blocks form the pipeline of the end-to-end system.}
    \label{fig:system_overview}
    \vspace{-0.5cm}
\end{figure}
In this section, we provide a high-level overview of ``Talk-to-Resolve (\system)''. It consists of four major components, as shown in Figure~\ref{fig:system_overview}.
\begin{enumerate}
    \item A \textit{task instruction understanding} component that parses a given instruction to predict the task type and arguments.
    \item A \textit{task (re)planning} component that generates the task execution plan by translating the parsed instruction to an abstract action sequence that the robot actuates to satisfy the task goal. It also re-plans after resolving visual ambiguity (if any) with the feedback through the dialogue.
    \item A component for \textit{visual uncertainty analysis} that locates the exact referred object(s) mentioned in the task instruction. In case there is visual uncertainty, it triggers the dialogue system to resolve the ambiguity by engaging with the human. If the referred object(s) can be visually pinpointed, the task execution continues; otherwise the task is aborted (after dialogue exchange with the human).
    \item A \textit{dialogue} system that maps the visual uncertainty to an ambiguity state and generates a question that suitably elucidates the ambiguity that the robot is facing.
\end{enumerate}
In this article, we focus on the visual uncertainty analysis and dialogue to resolve the uncertainty, where we incrementally ground the entities of the abstract plan. We attempt to visually ground only those entities that are required by the current action. During visual grounding, \system analyses the current ego-view of the robot to decide if such a grounding is possible. In the cases of grounding failure and ambiguity, it invokes the dialogue component, which uses visual uncertainty analysis to formulate questions for the human co-worker. As \system generates specific questions that can be answered by either a binary yes/no or choosing an answer from a set of suggestions, the answer is direct and unambiguous. In the case of an indirect answer, it is treated as a rephrased description of the same argument and processed in the same pipeline.
\section{Talk-to-Resolve (\system) in Details}
In this section, we describe our approach in detail and explain the integration of individual modules.

\subsection{Task instruction understanding}
We assume that a given instruction contains a task, having an unambiguous goal. The task is decomposed into a sequence of robot executable actions by task planning. To generate such a plan from the instruction, we utilize Conditional Random Field (CRF) models introduced in our previous work~\cite{pramanick2019enabling, pramanick2020decomplex}. The CRF models are trained to recognize a set of generic tasks with the corresponding arguments from text input, as shown in Table~\ref{tab:task-description}. The task types and arguments are defined according to the theory of semantic frames~\cite{baker1998berkeley}.
\begin{table}[t]
\caption{Description of task types that are used to train and test our system. \textit{Italicized} words are argument types associated with the task. The system is scalable and easily extendable for our types of tasks.}
    \centering
    \begin{tabular}{|c|l|}
    \hline
    \textbf{Task type} & \textbf{Description} \\ \hline
    Bringing & Bring an \textit{object} to a \textit{beneficiary} from a \textit{source}. \\ \hline
    Change\_state & Manipulate a \textit{device} to a desired \textit{state}. \\ \hline
    Check\_state & Assert the given \textit{state} of an \textit{object}. \\ \hline
    Motion & Move to a given \textit{goal}. \\ \hline
    Placing & Place an \textit{object} on a \textit{goal}. \\ \hline
    Searching & Look for an \textit{object} in an \textit{area}. \\ \hline
    Taking & Pick up an \textit{object} from a \textit{source}. \\
    \hline
    \end{tabular}
    \label{tab:task-description}
\end{table}
For example, the following instruction is annotated with the task and argument types by the task-CRF and argument-CRF models~\cite{pramanick2019enabling}, respectively.

\textit{\small [Take]\textsubscript{taking}  [a red cup]\textsubscript{object}  [from the kitchen]\textsubscript{source}}.

Given the predicted task type and the complete set of arguments for the task, we attempt to ground the arguments to the entities in the environment. In the example above, we consider grounding the argument values for \textit{object} and \textit{source} to a unique red-colored cup that is visible and a unique location called the kitchen, respectively.

Considering a robot in an office environment, these task types are chosen. However, the system design is scalable and easily extendable for other task types. This just requires the CRF models to be trained with datasets (annotated textual instructions) for those data types. Since the primary objective of this article is to resolve various ambiguity levels to ground the referred object, it is not dependent on the task type. Hence, the core contribution of this article remains applicable to the extended system as well.

\subsection{Plan generation}
Given the task type and arguments mentioned in the instruction, we generate a high-level task plan to execute the task. For each task type, we define a set of pre-condition and post-condition templates~\cite{pramanick2019enabling}. The templates are populated by the task and argument prediction and thus we encode the instruction into a PDDL planning problem~\cite{mcdermott1998pddl}. Finally, we use a forward-search planner to generate the task plan, i.e., the sequence of actions the robot needs to perform. An example of a task plan is given below.
\begin{enumerate}
    \item MOVE\_TO \textit{source}
    \item LOCALIZE \textit{object} \textit{source}
    \item PICK\_UP \textit{object}
    \item MOVE\_TO \textit{destination}
\end{enumerate}
We utilize the abstract task plan to enable the incremental grounding process. Before executing an action in the plan, such as \textit{MOVE\_TO}, we attempt to ground its argument(s), e.g., \textit{source} in the action. The \textit{source} location `kitchen', being a static geo-fencing area, the navigation goal can be determined from the robot's knowledge. Therefore, the execution of subsequent actions, where the object (a red cup) must be grounded visually, can be deferred until the robot reaches the source location. We assume that an initial occupancy map of the environment is known and the geo-fencing areas are annotated with names. For the entities that are not present in the knowledge base (such as movable objects), we resort to visual grounding.

\subsection{Visual uncertainty analysis}
\label{sec:visual-uncertanity}
The visual uncertainty analysis module aims to localize an argument to the bounding box of a unique entity in the scene. It also decides if this localization is uncertain and infers the nature of the uncertainty. To perform this localization, we first predict the entities present in the scene along with their bounding boxes and generate a description for each of them. To generate the descriptions in natural language, we utilize a dense image captioning network, \densecap~\cite{johnson2016densecap}. Given an image, it generates multiple region proposals, encodes the region features using a CNN, and uses a recurrent network (LSTM) to generate descriptions of the proposed regions. As the region proposal network in \densecap is not constrained to any particular object type, the generated descriptions are not restricted to the objects mentioned in the instruction. Therefore, to find the most likely candidate object(s) in the scene, we rank the generated captions according to their semantic similarity with the argument phrase. 

However, a direct semantic mapping is often not possible, as the perception of the same object can differ for the robot and a human due to poor lighting, viewing angle, clutter, and partial occlusion. Also, there can be multiple objects of the same type and due to mismatch in the vocabulary of natural language and that of the LSTM language model of \densecap, the same object can be referred to by different words. For example, in Figure~\ref{fig:representational}, there are three bottles, one containing wine and the others containing water. An instruction to simply fetch a bottle in this scenario is ambiguous. Also, an instruction like ``bring me a coffee mug'' does not have any lexical match with the description ``a red plastic cup'', although it is likely to be the intended object. In the same scene, the cup's material is also wrongly predicted to be plastic. Although, utilizing pre-trained word embeddings to compute the cosine distance between a pair of words~\cite{pennington2014glove} yields a semantic similarity metric that addresses some of these issues, defining such a similarity function to compare a caption and an argument is still non-trivial.

\begin{itemize}
    \item Both the argument and the caption contain a variable number of tokens, so the pairwise similarity between the tokens~\cite{pennington2014glove} cannot be calculated directly.
    \item Although vector-composition based models~\cite{arora2017simple,cer2018universal} encode a variable-length phrase, they do not yield an optimal relevancy ranking. This is because such models aim to capture a generic summary of the word embeddings, failing to exploit important local features. For example, if the instruction is ``turn off the red lamp'', a caption \textit{`a lamp on the table'} is relevant than \textit{`a red table'}.
    \item The generated captions are complete sentences. Whereas the argument is often an incomplete phrase, which leads to a large length mismatch. This limits the applicability of n-gram alignment techniques such as METEOR~\cite{banerjee2005meteor} used in~\cite{shridhar18}.
\end{itemize}

Therefore, we introduce a new semantic similarity metric, where we encode the words into pre-trained GloVe embeddings~\cite{pennington2014glove} and compute a convex combination of the embeddings to generate a vector of fixed dimension. In contrast to existing vector composition models that assign weights to individual word vectors~\cite{iyyer2015deep,arora2017simple}, we classify the words into a set of semantic classes and only assign weights to the semantic classes. This results in a much simpler learning problem, as we only optimize weights for $k$ semantic classes, as opposed to learning a composite, $d$ dimensional embedding $(k<<d)$, through supervision~\cite{iyyer2015deep}. In the following, we define the similarity metric and describe how it is applied to determine an ambiguity state for question generation.

\subsubsection{Semantic similarity calculation}
The \densecap model outputs a set of bounding boxes $B_{1:n}$ and captions $C_{1:n}$ for a given scene. We define the semantic similarity function $f(A,c_i)$, to compare the argument phrase $A$ with a given caption $c_i$, and thus find an optimal relevancy ranking. Firstly, given a sequence of tokens, we predict a semantic class $s \in SC$ for each token. In our context, $SC$ consists of the classes - \textit{object}, \textit{attribute}, \textit{spatial\_landmark} and a \textit{others} class to account for any other class of word. We model this semantic class prediction as a sequence labeling problem and train a CRF model to perform the labeling.
Given a token sequence $t_{1:n}$, we perform inference on the CRF to find the most probable sequence of semantic class labels $s_{1:n}$,
\[  s_{1:n} =\argmax_{s_i \in SC} P(s_i | t_{1:n},s_{i-1})  .\]
We extract several grammatical features in the feature functions of the CRF. The features include the lemma, POS tag, and dependency tag of the token and its direct neighbors of the dependency parse tree, including the previous prediction $s_{i-1}$ as the transition feature. To estimate a composite weighted embedding of a sequence of tokens $t_{1:n}$, we interpolate the token embeddings with the corresponding class weight. Therefore, given $n$ token embeddings of an argument phrase or a caption as $d$ dimensional vectors, we find the convex combination as,
\[ V^d = \sum_{i=1}^n{\lambda_i E(t_i)^d},\]
where $\lambda_i$ and $ E(t_i)$ are the interpolation weight and the embedding of the token $t_i$, respectively. Given the encoded argument as $V^d_A$, and an encoded caption as $V^d_{c_i}$, we find their pairwise similarity as,
\[ f(A,c_i) =  \frac  { V^d_A \cdot V^d_{c_i}} {\| V^d_A\| \|V^d_{c_i}\|} .\]

\subsubsection{Redundancy suppression}
Given the set of bounding boxes and the corresponding captions, ranked according to the caption's similarity with the argument, we analyze the bounding boxes for possible redundancy. The caption generation model predicts the most probable sequence of words, given an image region as a proposal. However, such region proposals often overlap with each other, which results in redundant caption generation. Although a greedy non-maximum suppression (NMS) can be applied to prune region proposals with high overlap~\cite{johnson2016densecap}, \ie Intersection over Union (IoU), setting the IoU threshold is generally difficult~\cite{bodla2017soft}. The difficulty increases when the robot must detect objects of varying size, distributed in varying distances; where larger and closer objects may lead to multiple captions. Thus naive NMS often fails to suppress captions that are about the same object, having a slightly different sequence of words. We consider two distinct types of redundancy in the generated captions and tackle them with different strategies.
 \begin{itemize}
     \item Object redundancy - when multiple bounding boxes are proposed for the same object that results in captions, where either no attribute is associated with the object, or the attributes are the same across the captions.
     \item Caption redundancy - when multiple captions are generated for the same object, whose attribute sets are disjoint. 
 \end{itemize}
Resolving object redundancy is important for avoiding the false detection of ambiguity. Whereas, in the case of actual ambiguity, multiple instances of the same object must be considered separately. In the case of caption redundancy, although redundant captions are suppressed, the distinct attributes are merged to capture a distinct description of the object. We apply a greedy heuristic to keep the most relevant captions by jointly applying a semantic similarity and an IoU cutoff. Given the ranked captions, $c_{i:n}$, we consider $c_1$ to be the most probable candidate, \ie 
\[c_1 = \argmax f(A,c_i)\]

Firstly, we prune irrelevant captions by applying a semantic similarity cutoff $\alpha$. Thus, we estimate the set of relevant captions,

\[ C_R = \{c_i : f(A,c_i) > \alpha, 1<i\leq n\}\].

Then we find the set of candidate captions by pruning the remaining captions $c_i \in C_R$ that satisfy the following,

\[ IoU(c_i,c_1) > \beta , f(c_i,c_1) > \alpha ,\]

where $\alpha$ and $\beta$ are the semantic similarity and IoU cutoffs, respectively. While pruning a relevant caption, \ie $c_i \in C_R$, we suppress caption redundancy by utilizing the previously predicated semantic class labels. Let $t_o$ denote the token corresponding to an \textit{object} class label in $c_1$. Then if $t_o$ is present in a caption to be pruned, we merge the tokens in $c_i$ having the \textit{attribute} label, with the set of attribute tokens in $c_1$. While merging such a caption $c_i$ with $c_1$, we also change the bounding box of $c_1$ as $b_1'=b_1 \cup b_i$. We find the optimal values of $\alpha$ and $\beta$ through a grid search in the range $(0,1)$ minimizing error in ambiguity state identification, using a validation dataset.

\subsection{Dialogue system}
\begin{figure}
    \centering
    \includegraphics[width=\linewidth]{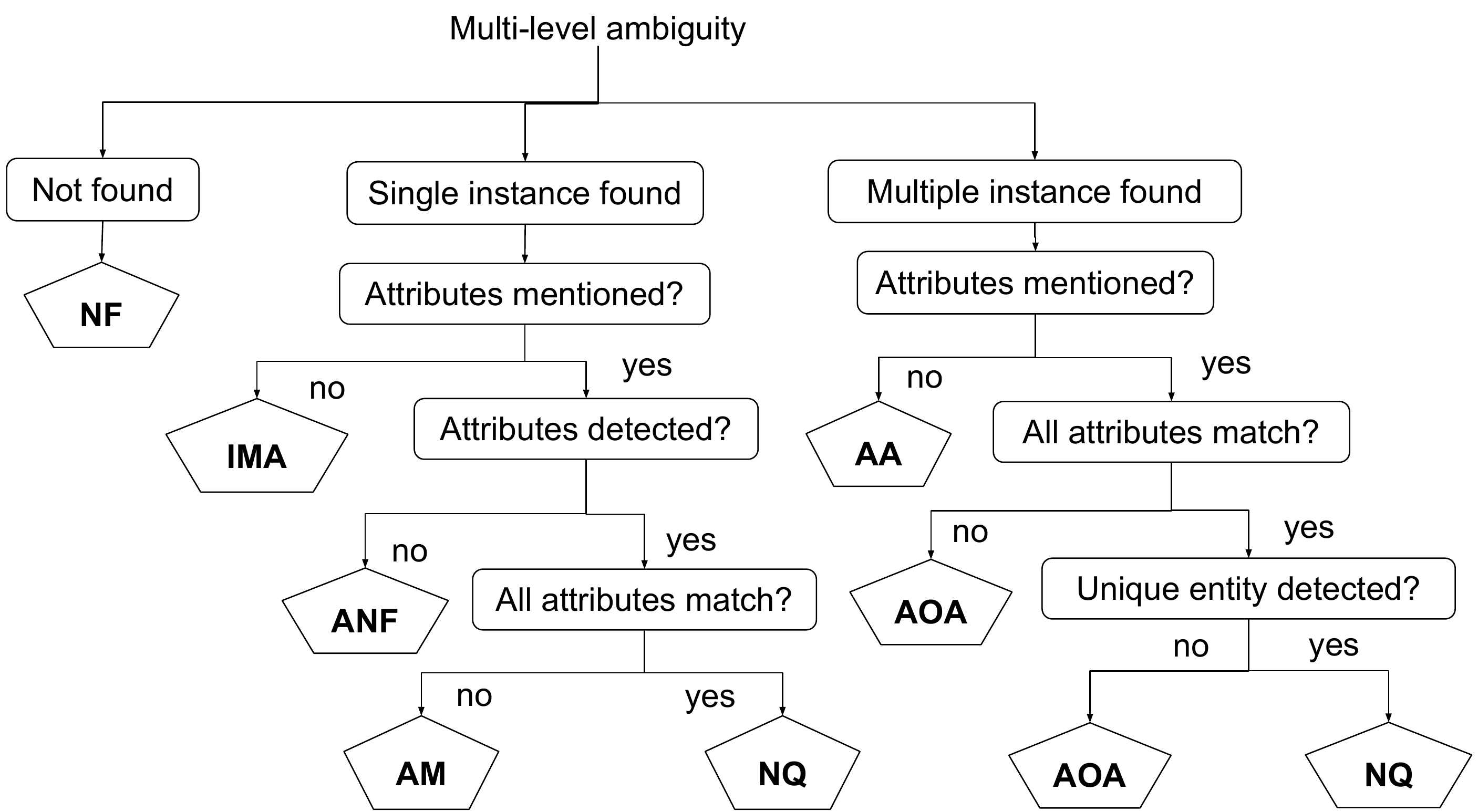}
    \caption{Deriving the multi-level ambiguity states (listed in Table~\ref{tab:state-description}) considering all possible ambiguous scenarios.}
    \label{fig:multi-level-ambiguity}
\end{figure}

We analyze the final set of candidate captions to decide if a question should be asked. We define a set of ambiguous states so that the specific problem faced by the robot is mapped to one of the states, and an appropriate question for that state can be formed. Unlike the existing work that decides only binary ambiguity, i.e., there is ambiguity only if there is more than one object, we define multi-level ambiguity states. Firstly, ambiguity can arise even if there is only one object -- the physical or spatial properties mentioned in the instruction do not match with the object on the ground. Secondly, ambiguity can arise if there are multiple instances of the objects along with the object properties that may or may not match with the description (if any) in the instruction. Sometimes, even if there are multiple instances of the referred object, which is usually termed as ambiguity in most of the existing works, it may not be ambiguous if only one instance matches with the object description in the instruction. Our system handles all such scenarios, which leads to a multi-level ambiguity resolution. As a result, the query generated by the system is more informative and contextual, which helps the human to realize the exact nature of the impasse that the system is facing. Based on our analysis of various situations and instruction, we define seven such states, as described in Table~\ref{tab:state-description} that are sufficient to capture all possible object ambiguities. The analysis is depicted in Figure~\ref{fig:multi-level-ambiguity} considering the possible scenarios that may arise on the ground. In the following, we describe our approach to detect the ambiguity state and generate the question given the detected state.
\begin{table}
    \centering
    \caption{Description of ambiguity states identified by comparing visual information with the argument.}
    \begin{tabular}{|p{3.9cm}|p{8.6cm}|}
    \hline
        State & Description \\ \hline
        No question (NQ) & All the information is available. \\ \hline
        Ambiguous \newline attribute (AA) & Multiple matching objects, but no attribute mentioned in instruction. \\ \hline
        Implicitly matching attribute (IMA) & Unique object with attribute, but no attribute mentioned in instruction. \\ \hline
        Attribute \newline mismatch (AM) & Unique object, but its attribute is different from the instruction. \\ \hline
        Attribute not found (ANF) & Unique object without attribute, but attribute is mentioned in instruction. \\ \hline
        Ambiguous object and attribute (AOA) & Multiple matching objects that have either none or the same attributes. \\ \hline
        Not found (NF) & The object can't be found, possibly an error in object detection. \\ \hline
    \end{tabular}
    \label{tab:state-description}
\end{table}
\subsubsection{Ambiguity state identification}

To identify a dialogue state we again utilize the CRF model we introduced for semantic class prediction. We extract the semantic entities from the argument and compare them with extracted entities in the candidate captions, utilizing the semantic classes that are predicted in the visual uncertainty analysis stage (Section~\ref{sec:visual-uncertanity}). Specifically, we check if the same object token is mentioned in the candidate captions. If there is a unique candidate with a matching object, we check the attribute tokens of both the argument and the candidate. In the case of an argument mismatch, the AM state is identified, and ANF is identified when the argument is missing in the candidate. Also, we take the strategy to confirm the object's attribute if not explicitly mentioned in the argument, i.e the IMA state. Otherwise, in the case of a unique and exact match, we refrain from asking questions (NQ).

In the case of multiple candidates, we check if any attribute is mentioned in the argument. If so, we check if multiple candidates with matching objects have either the same or no attribute and thereby identify the AOA state. Otherwise, we ask for clarity on the attribute for disambiguation, therefore identifying the AA state. Finally, if no candidates are found, or there is no matching object in the set of candidates, we decide on the NF state.

\subsection{Question generation}
\system generates a question if the detected state is the one where an argument can't be uniquely grounded, or such a grounding is uncertain. The question is crafted to convey the robot's partial understanding of the scene and also pinpoint the ambiguity/uncertainty. To generate such questions, we use a set of templates, where each template contains multiple slots, that are filled using the semantic entities and filler words, from both the instruction and the candidate captions. We designed the question templates in a pragmatically appropriate manner and present the user with choices whenever applicable. Moreover, we left out any task-specific terms in the templates so that they are generalized across multiple task types. The templates used in \system are shown in Table~\ref{tab:question-templates}. The question templates specifically mention the mismatch for single object instances so that a binary answer can be provided to continue the grounding. In the cases of multiple object instances, the discriminative attributes are mentioned in the question such that the user can choose one, except for the AOA state. In AOA, there are multiple visually similar objects (e.g., multiple green bottles on a table) and therefore the user must specify some discriminating attribute in the response or allow the robot to choose the physically closest object.

The question generator uses the predicted semantic classes of the argument and the candidate captions to replace the occurrence of the corresponding slots. Although we define one template per ambiguity state for our experiment, it is possible to use multiple templates for the same state, making the questions seem non-repetitive. This can be done by re-phrasing the template, utilizing the same set of slots.
\begin{table}
    \centering
    \caption{Mapping of question templates with ambiguity states, where the \underline{underlined} and the \textbf{boldface} slots are filled from the candidate captions and the argument, respectively.}
    \begin{tabular}{|c|p{11.5cm}|}
    \hline
    State & Template \\ \hline
    AA & I see a \underline{attribute-1} \textbf{object} and a \underline{attribute-2} \textbf{object}. Which one did you mean? \\ \hline
    ANF & I see a \textbf{object}, but not sure if it’s \textbf{attribute}. Should I continue? \\ \hline
    IMA & I see a \underline{attribute} \textbf{object}. Should I continue? \\ \hline
    AM & I see a \textbf{object}, but its \underline{attribute}. Should I continue? \\ \hline
    AOA & I see \underline{\#num} \textbf{attribute} \textbf{object}s. Which one did you mean? \\ \hline
    NF & I can't find any \textbf{attribute} \textbf{object}. What should I do? \\ \hline
    \end{tabular}
    \label{tab:question-templates}
\end{table}
\section{Evaluation}
We evaluate our system using a curated dataset of indoor scenes, task instruction, and appropriate ambiguity state triplets. We have introduced a CRF model for semantic class prediction as a crucial component of our system. Therefore, we also evaluate the model using a separate dataset of annotated semantic class labels.


\subsection{Datasets}
To train the CRF for semantic class prediction, we have collected a total of 242 object descriptions from the Visual Genome dataset~\cite{krishna2017visual} that is used to train~\densecap~\cite{johnson2016densecap}. We have sampled descriptions of image regions around everyday objects from 40 different indoor images. We have tokenized and annotated each token of a description as a sequence of semantic class labels, using a text annotation tool~\cite{yang-etal-2018-yedda}. This results in the annotation of 4.48 (\textit{SD=1.19}) classes per description, on average. We have trained the CRF with 80\% of the data, and have evaluated using the remaining 20\%.

To evaluate our ambiguity state identification method, we have collected a total of 88 indoor scenes from an indoor scene recognition dataset~\cite{quattoni2009recognizing}. For each image, we have written multiple instructions conveying different task types and referring to different objects in the image. Also for every object type in an image, we have written instructions by referring to the same object type with a varying granularity of attributes and intentionally mistaken attributes. This results in a balanced dataset of different ambiguity and mismatch scenarios. We annotated the most appropriate ambiguity state for a given image-instruction pair. Two of the authors have written the instructions and annotated the states. Another author reviewed and corrected the annotations for the entire dataset. The final dataset contains 358 image-instruction pairs. There are 7 different task types and 7 different states in the dataset, as shown in Table~\ref{tab:task-description} and Table~\ref{tab:state-description}, respectively. We select a random split of 10\% of the data as a validation set for tuning the semantic class weights $\lambda_i$, the semantic similarity, and IoU cutoffs and use the remaining 322 image-instruction pairs as the test set.

\subsection{Performance of semantic class prediction}
We compare our CRF model with a grammar-based baseline, where we convert a dependency parse tree of the text to semantic class labels. Following the parse tag definitions in~\cite{nivre2016universal}, we label a token as the \textit{Object} class if it is the root of the parse tree. Subsequently, we label the tokens having adverbial modifier (`amod'~\cite{nivre2016universal}) dependency as \textit{Attribute} and having the indirect object dependency (`iobj'~\cite{nivre2016universal}) as \textit{Spatial\_landmark}, while any other tag is labeled as \textit{Other} class. The semantic class prediction results are shown in Table~\ref{tab:semantic-results}. The results suggest that the CRF surpasses the grammar baseline and has decent accuracy for use in the semantic similarity function and caption comparison. 
\begin{table}
    \centering
    \caption{Semantic class labeling result (F1-score) of our CRF model, compared to a grammar based baseline.}
    \begin{tabular}{|l|c|c|}
    \hline
    Semantic class & Grammar\textsuperscript{ baseline} & CRF\textsuperscript{ ours}\\ \hline
    Object &0.76 & \textbf{0.91} \\ 
    Attribute & 0.67 & \textbf{0.96} \\
    Spatial\_landmark & 0.83 & \textbf{0.91}  \\
    Other & 0.74 & \textbf{0.97} \\ \hline
    Avg. & 0.72 & \textbf{0.95} \\  \hline
    \end{tabular}
    \label{tab:semantic-results}
\end{table}

\subsection{Performance of ambiguity state identification}
We compare our approach with three baseline systems, where we analyze different semantic similarity metrics for caption relevancy ranking and measure the effect of redundancy suppression. The following describes the system variants used in the experiment.
\begin{itemize}
    \item METEOR - We use the METEOR metric~\cite{banerjee2005meteor} as the semantic similarity function to prune irrelevant captions, but do not suppress redundancy.
    \item Deep Averaging Network (DAN) - We use a pre-trained deep averaging network~\cite{cer2018universal} to encode the caption and the argument, followed by a cosine similarity computation. No redundancy suppression is applied to the ranked captions.
    \item S-BERT - We use a transformer-based network~\cite{reimers-2019-sentence-bert} to directly compute the caption and argument embeddings for cosine similarity computation, without redundancy suppression. Specifically, we use the \textit{paraphrase-albert-small-v2} model that was trained for a paraphrasing task, which has a similar objective to ours.
    \item Semantic Weight Interpolation (SWI) - Our proposed weighted vector composition model is used for semantic similarity, without redundancy suppression.
    \item \system - Our full model, where both the object and caption redundancy suppression are applied, along with our proposed vector composition model for semantic similarity.
\end{itemize}
We optimize the cutoff thresholds for all the baseline systems using the same validation set and use the same task and argument labeling models. Also, we augment the proposed CRF model for semantic class labeling in all the baselines to enable the state prediction.  By comparing with the test data annotation, we report the results in Table~\ref{tab:state-result}.
\begin{table}
    \centering
    \small
    \caption{Ambiguity state identification results (F1 score) of \system and baselines. Boldface numbers are highest.}
    \begin{tabular}{|c|c|c|c|c|c|}
    \hline
     State & METEOR\textsuperscript{ baseline} & DAN\textsuperscript{ baseline} & S-BERT\textsuperscript{ baseline} & SWI\textsuperscript{ ours} & TTR\textsuperscript{ ours}  \\ \hline
         AA     &\textbf{0.79}   &0.49  &0.70    &0.73  &0.76 \\
         IMA    &0.64   &0.74   &0.79    &0.82  &\textbf{0.85} \\
         AM     &0.60   &0.62   &0.68    &\textbf{0.73}  &\textbf{0.73} \\
         ANF    &0.42   &0.56   &0.73      &\textbf{0.75}  &\textbf{0.75} \\
         AOA    &0.58   &0.58  & 0.64   &0.60  &\textbf{0.65} \\
          NF    &0.84   &0.88   & 0.93    & \textbf{0.94}  & \textbf{0.94} \\
          NQ    &0.57   &0.58   & 0.65    &0.56  &\textbf{0.68} \\
      \hline
      Avg.      &0.71   &0.72   &0.77    &0.80  &\textbf{0.82} \\ \hline
    \end{tabular}
    \label{tab:state-result}
\end{table}

\subsubsection{Quantitative results}
The baseline system that uses METEOR, closely resembling the work of Shridhar and Hsu~\cite{shridhar18}, can only achieve an overall F1 score of 0.71. We observe that it is somewhat accurate in predicting the AA state, where the n-gram alignment in METEOR gives a better ranking to captions with a matching object type, whose attribute set is empty. However, it fails to tackle slight dissimilarities in the attributes, resulting in poor performance on the IMA, AM, and ANF states. For the prediction of these states, it is necessary to consider captions as relevant, even if the captions with a matching object have no attribute or have a different set of attributes. We see the opposite effect when using DAN, where the word vectors corresponding to the object are not explicitly given a high weight during composition, leading to low-ranking of captions describing the objects without any attribute, thereby failing to predict AA accurately. However, the continuous-space word representation of DAN slightly improves the accuracy for other states, improving the overall accuracy to 0.72. S-BERT largely outperforms DAN achieving an overall F1 score of 0.77 and also improving on all individual states. This shows that S-BERT produces a good, generic representation of the captions, being trained with a state-of-the-art transformer network and optimized for paraphrasing using multiple large datasets. 

Our proposed method of weighted vector interpolation (SWI) outperforms METEOR and DAN by large margins. We achieved a 0.8 score, even without redundancy suppression, particularly improving for IMA, AM, and ANF states. SWI also outperforms S-BERT in overall score, improving performance for most of the states. Finally, our full model achieved the best score of 0.82, also achieving the best scores in individual states, except for AA. Suppressing redundancy prominently helps in ambiguity identification, specifically when deciding between IMA vs. AA and AOA vs. NQ states. Please note that the detection of the NQ state is more difficult than other ambiguous states as the system needs to correctly localize and count the object instances, while also parsing the attributes from both the instruction and the captions perfectly (as shown in~\Cref{fig:multi-level-ambiguity}). In contrast, considering the IMA state as an example, simply detecting a single object instance that is mentioned in the instruction is assumed to be correct, regardless of any error in parsing the attributes of the caption describing the object. 

\subsubsection{Qualitative results}
\begin{figure}
     \centering
     \begin{subfigure}[b]{0.31\textwidth}
         \centering
         \includegraphics[width=\textwidth]{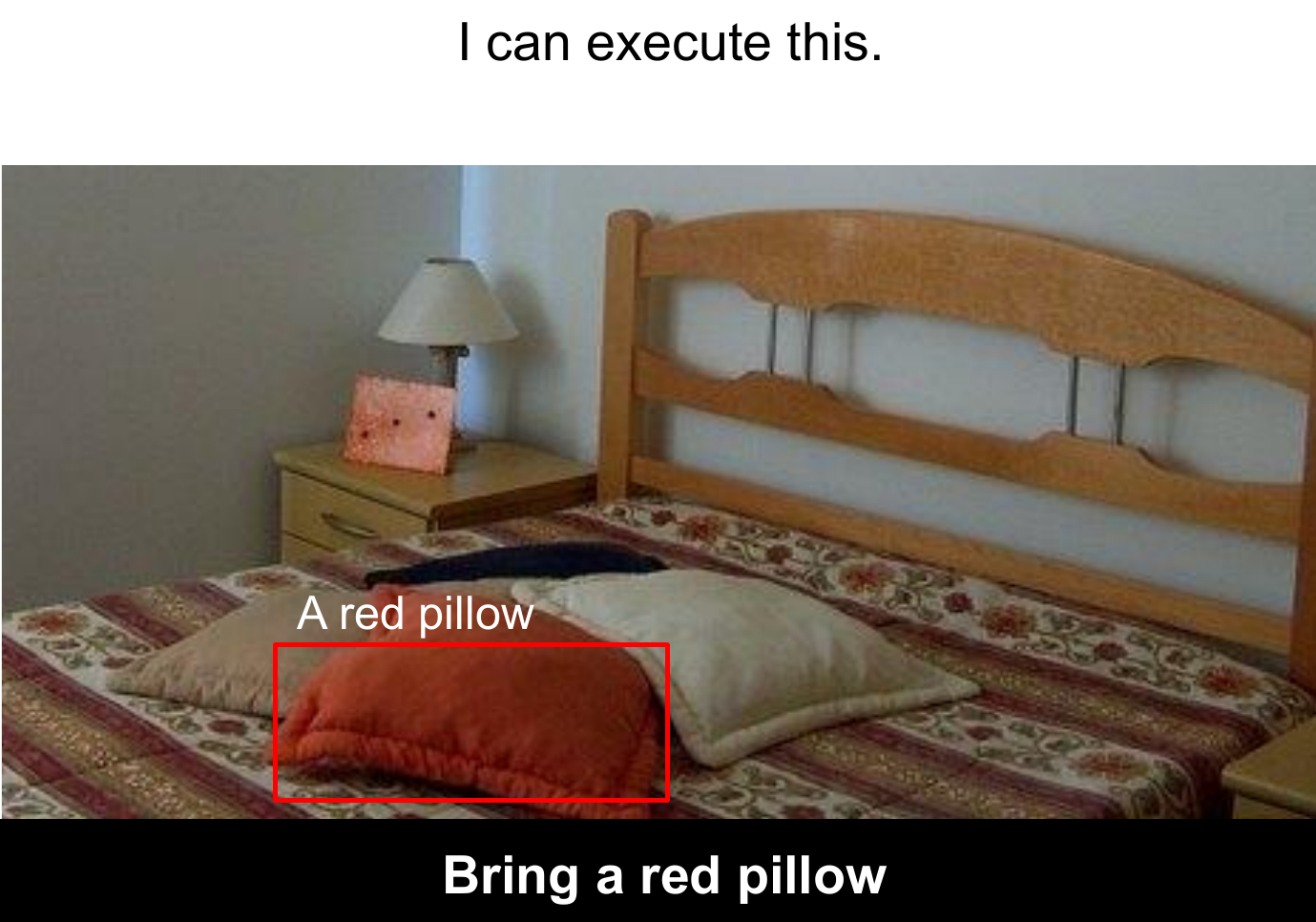}
         \caption{NQ}
         \label{fig:NQ}
     \end{subfigure}
      \hfill
     \begin{subfigure}[b]{0.31\textwidth}
         \centering
         \includegraphics[width=\textwidth]{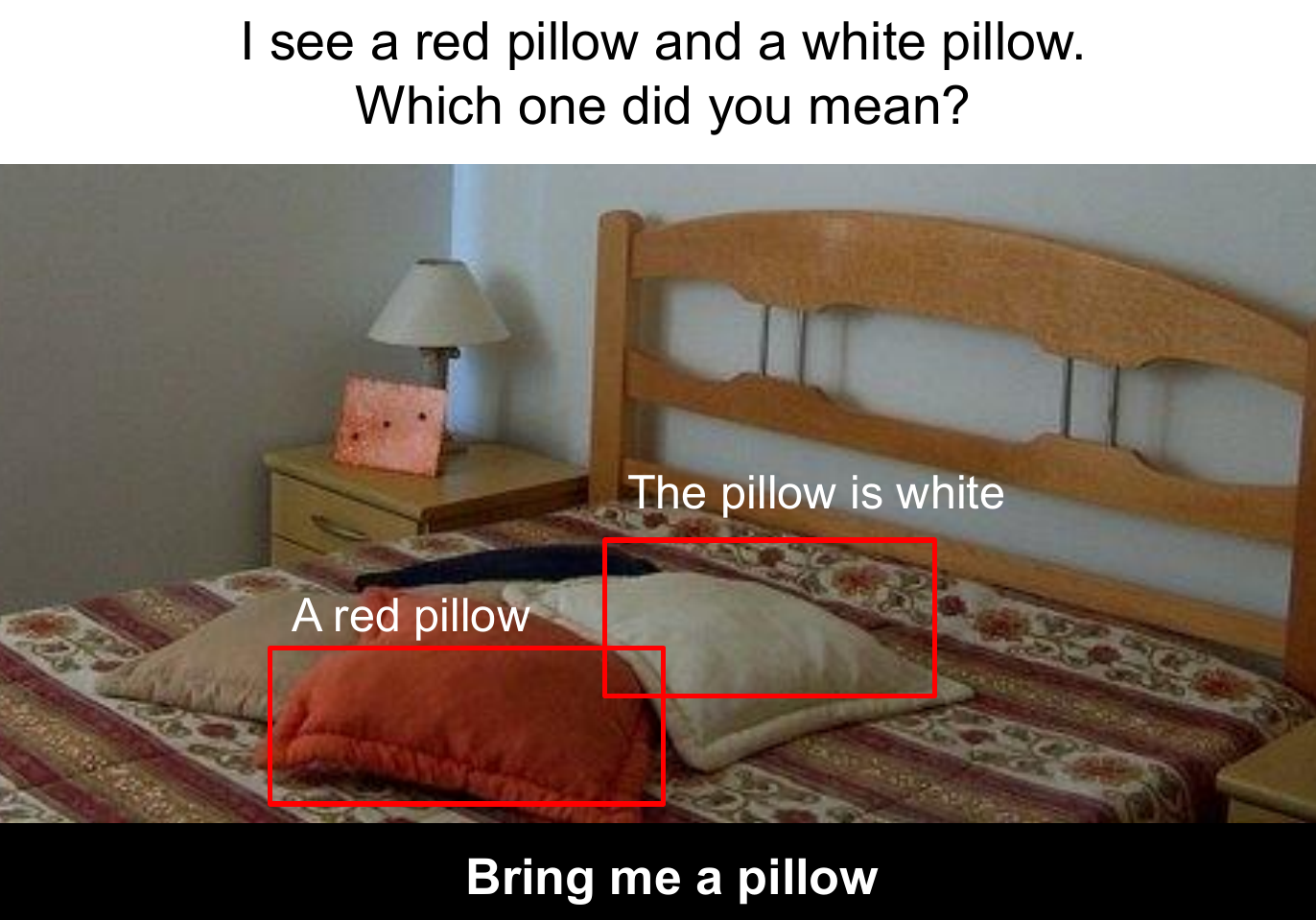}
         \caption{AA}
         \label{fig:AA}
     \end{subfigure}
     \hfill
     \begin{subfigure}[b]{0.31\textwidth}
         \centering
         \includegraphics[width=\textwidth]{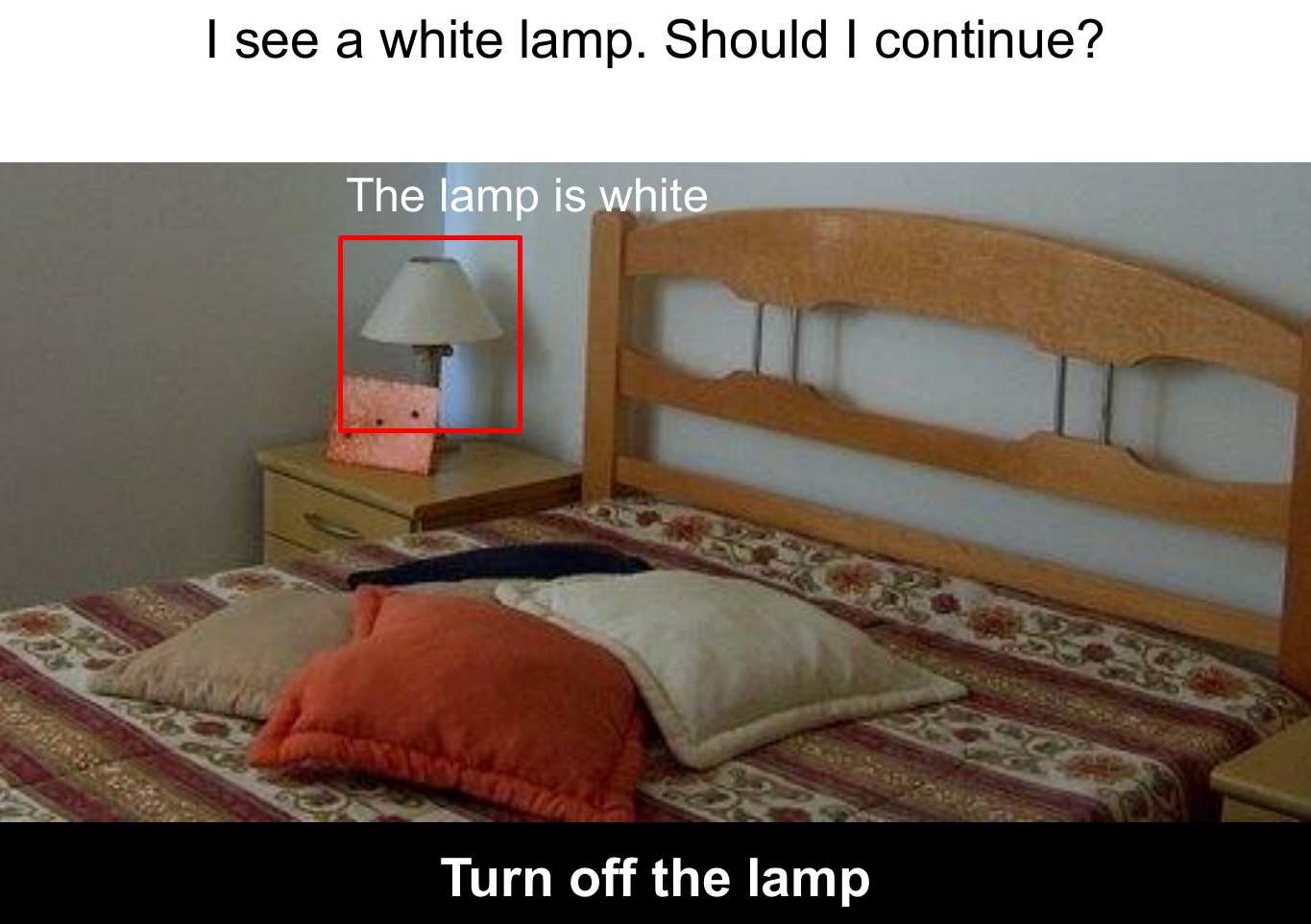}
         \caption{IMA}
         \label{fig:IMA}
     \end{subfigure}
     
     \begin{subfigure}[b]{0.31\textwidth}
         \centering
         \includegraphics[width=\textwidth]{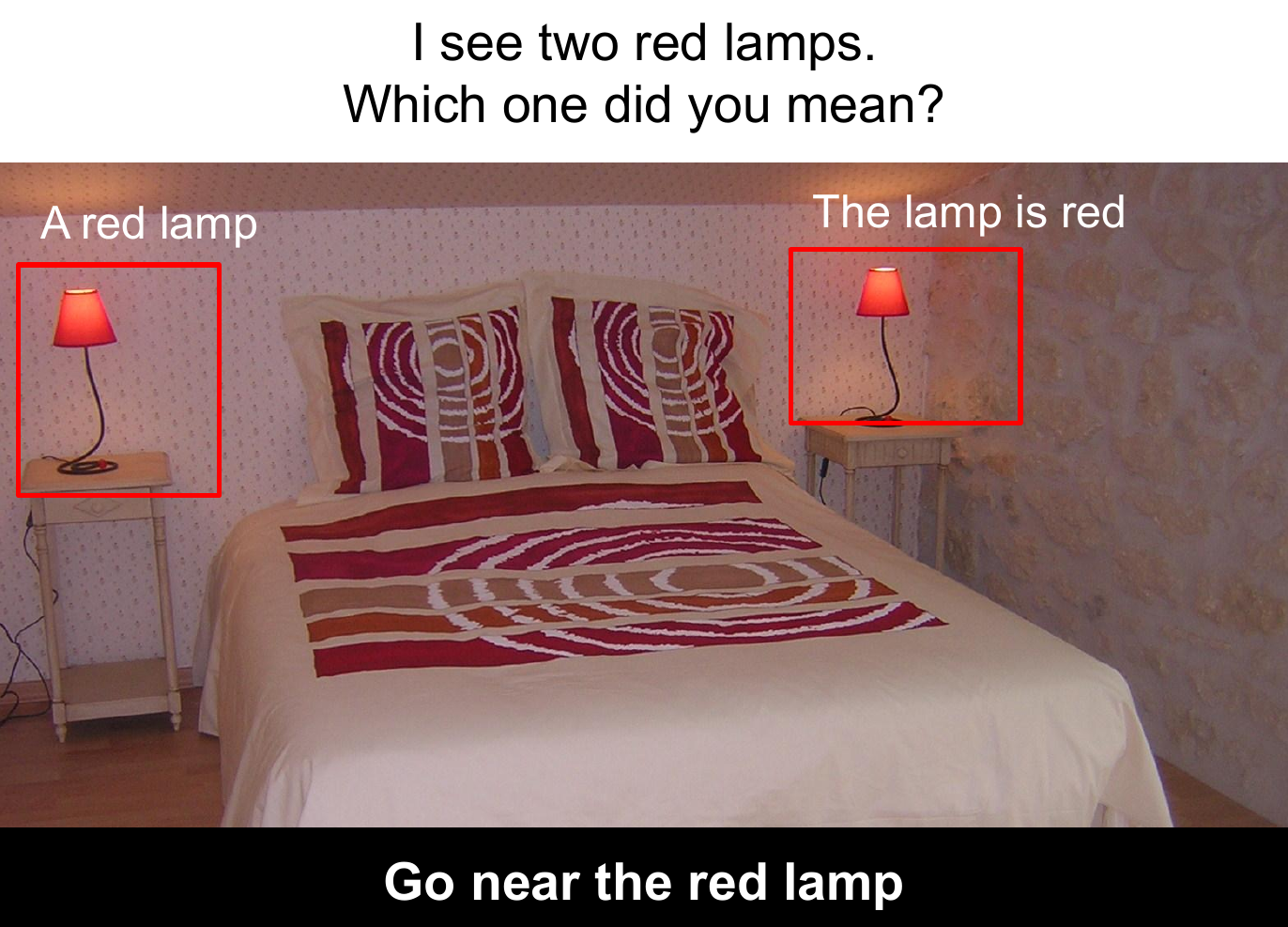}
         \caption{AOA}
         \label{fig:AOA}
     \end{subfigure}
     \hfill
     \begin{subfigure}[b]{0.31\textwidth}
         \centering
         \includegraphics[width=\textwidth]{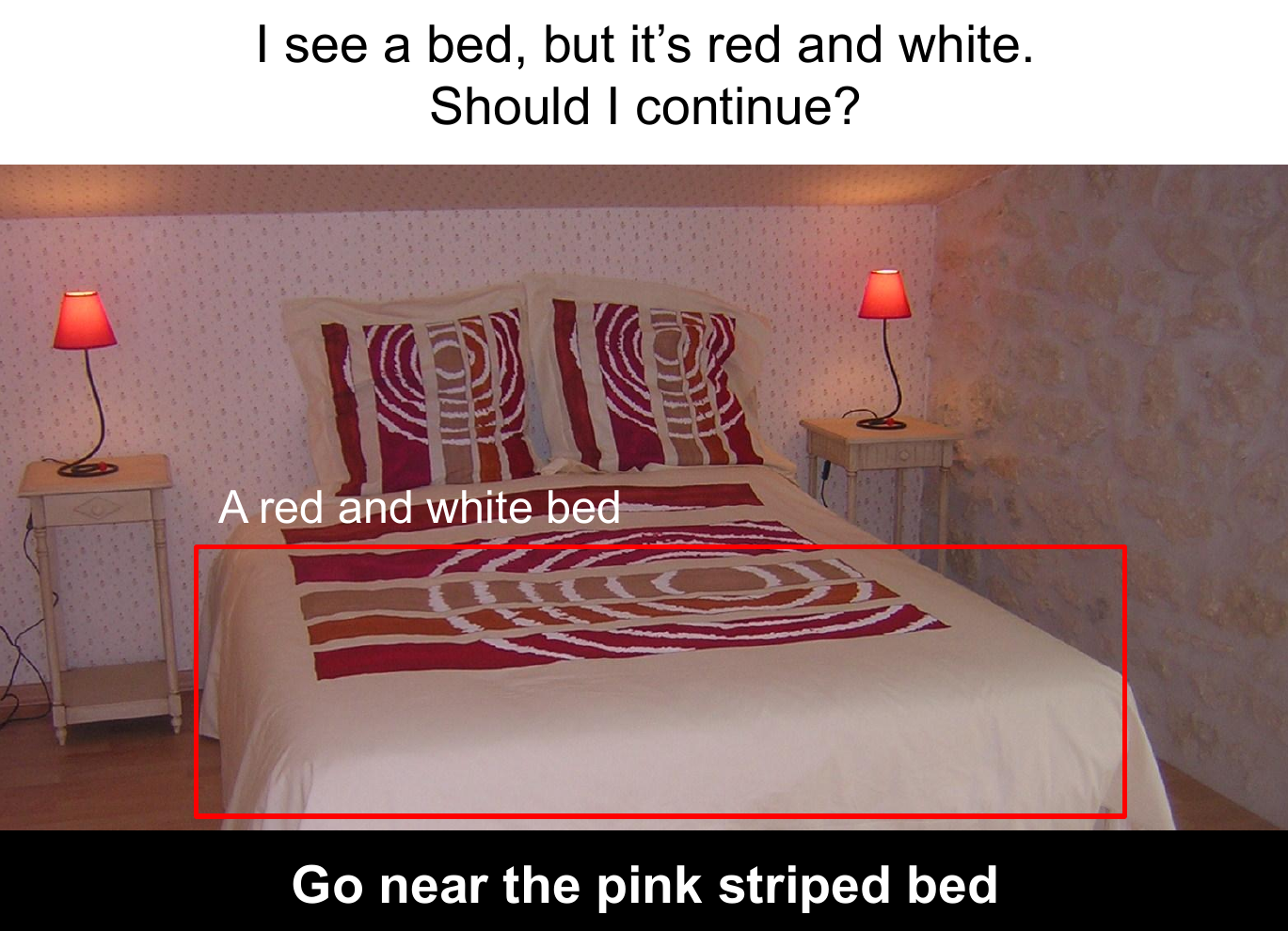}
         \caption{AM}
         \label{fig:AM}
     \end{subfigure}
     \hfill
     \begin{subfigure}[b]{0.31\textwidth}
         \centering
         \includegraphics[width=\textwidth]{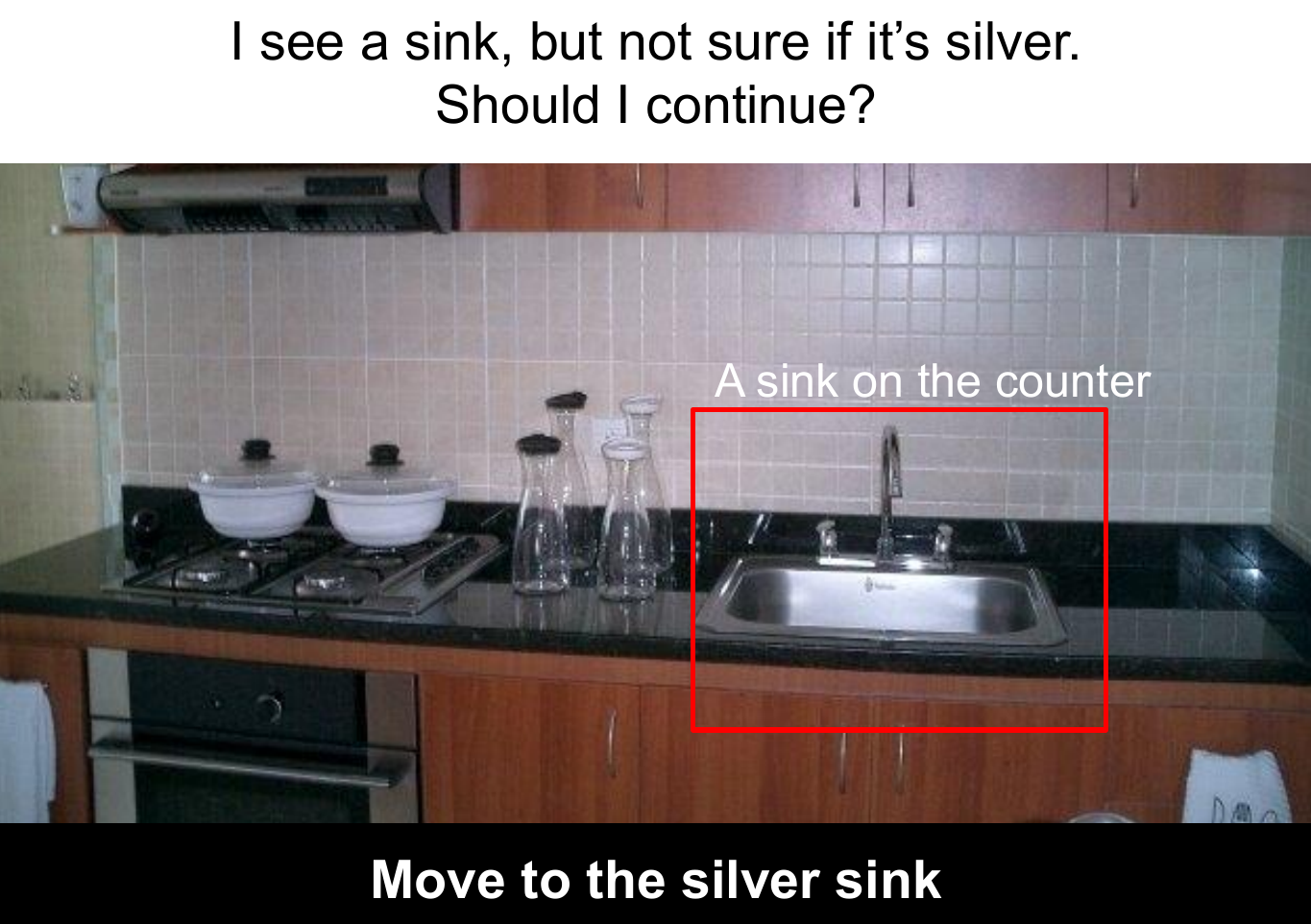}
         \caption{ANF}
         \label{fig:ANF}
     \end{subfigure}
        \caption{Examples of image-instruction pairs given to \system. The images are annotated with the final candidate captions selected by \system. The instructions are shown on the bottom and the responses generated by \system are shown on the top of each image.}
        \label{fig:qualitative-results}
\end{figure}

\begin{figure}
    \centering
    \begin{subfigure}[b]{0.31\textwidth}
         \centering
         \includegraphics[width=\textwidth]{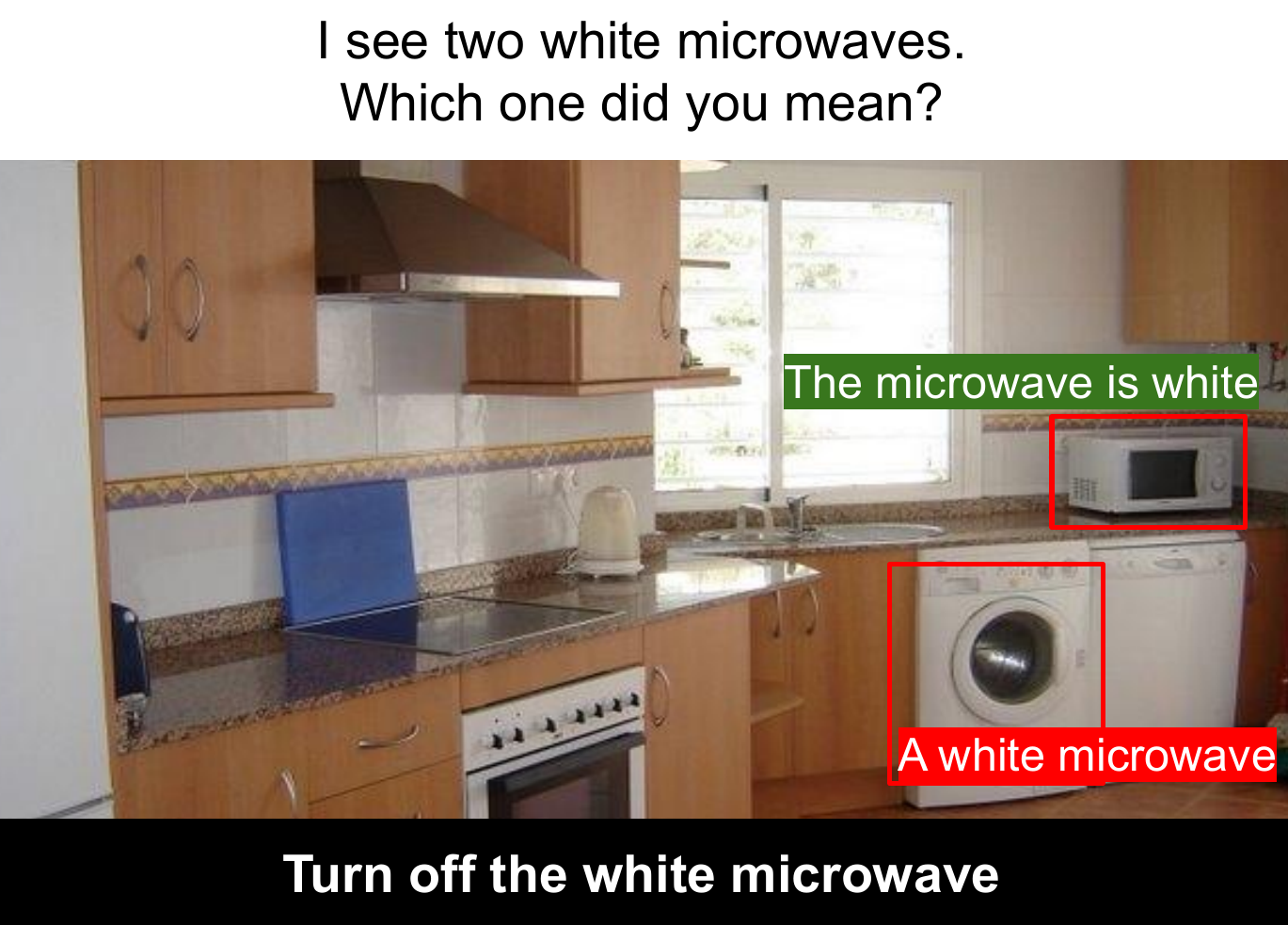}
         \caption{NQ, predicted AOA.}
         \label{fig:fail_AOA_NQ}
     \end{subfigure}
     \hfill
     \begin{subfigure}[b]{0.31\textwidth}
         \centering
         \includegraphics[width=\textwidth]{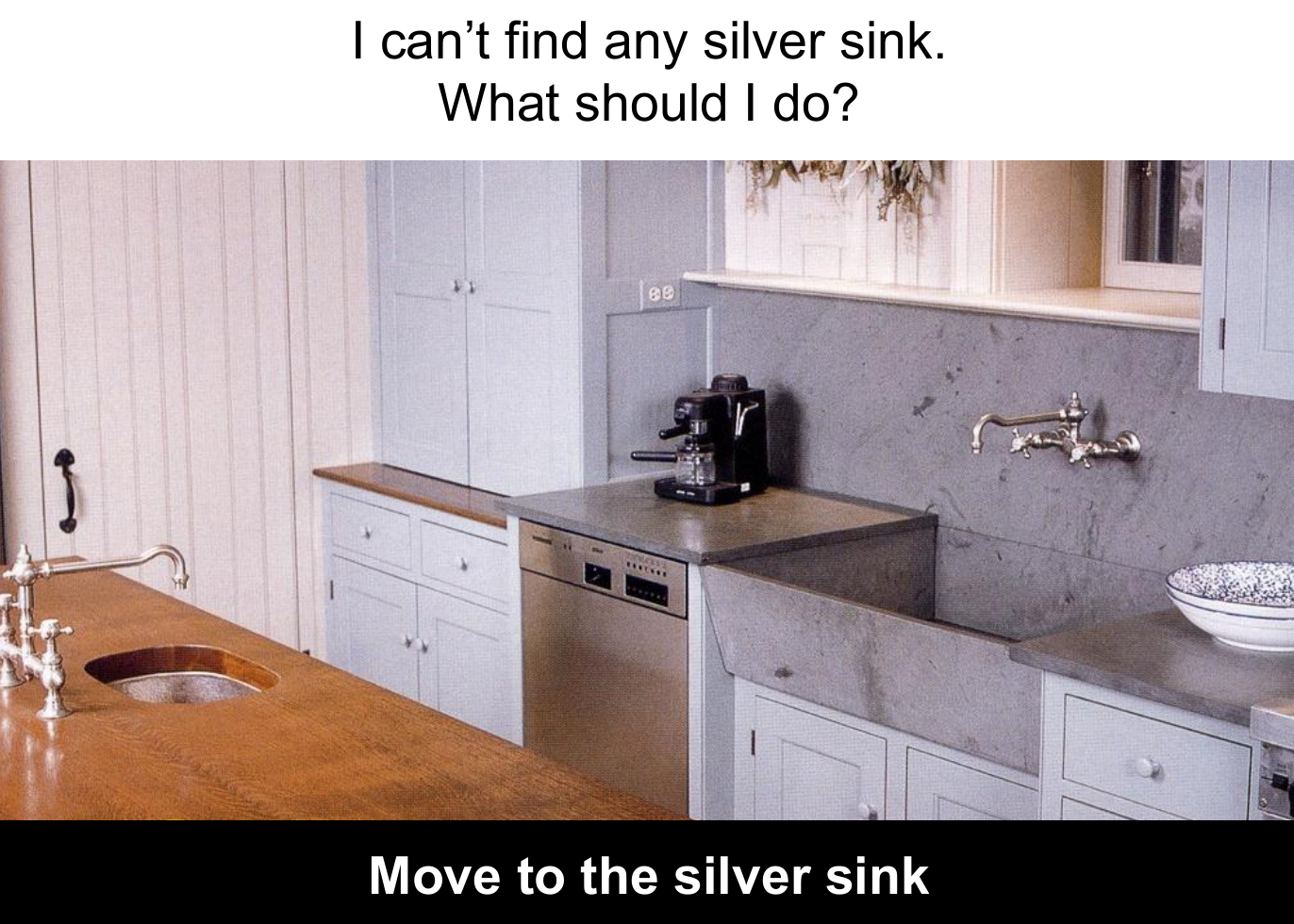}
         \caption{AM, predicted NF.}
         \label{fig:fail_NF_AM}
     \end{subfigure}
     \hfill
     \begin{subfigure}[b]{0.31\textwidth}
         \centering
         \includegraphics[width=\textwidth]{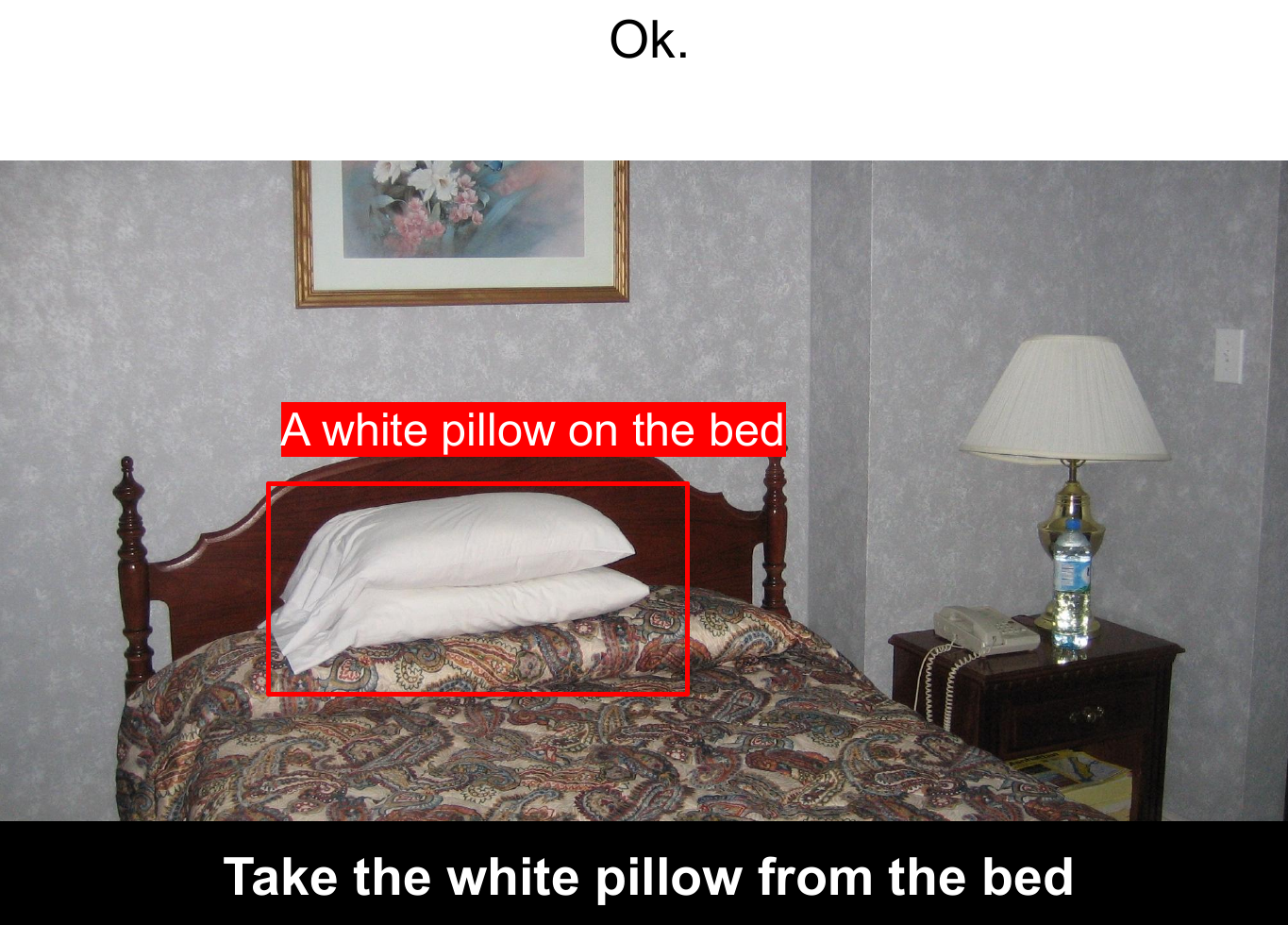}
         \caption{AOA, predicted NQ.}
         \label{fig:fail_NQ_AOA}
     \end{subfigure}
    \caption{Examples of incorrect ambiguity state prediction by \system. }
    \label{fig:faliure-cases}
\end{figure}

\Cref{fig:qualitative-results} shows some examples of image-instruction pairs given to \system and the corresponding ground truth ambiguity states, as defined in Table~\ref{tab:state-description}. As seen in~\Cref{fig:AA,fig:IMA,fig:AOA,fig:AM,fig:ANF}, the questions generated by \system are object-specific and clearly describe the problem in grounding. For the NQ state (\Cref{fig:NQ}), the acknowledging response asserts to the user that no further dialogue is required. Utilizing the semantic class prediction CRF, \system is able to consider lexically different captions as instances of the same object class, e.g., different captions of a lamp in \Cref{fig:AA}. It can also parse multi-token attributes such as `red and white' in \Cref{fig:AM}. 

\Cref{fig:faliure-cases} show some failure cases commonly encountered by \system. Problems in perception, such as mispredicting visually similar objects, result in false-positive detection of ambiguity. For example, in the scenario shown in~\Cref{fig:fail_AOA_NQ}, there should be no ambiguity regarding the single microwave present in the scene. However, as the washing machine is mispredicted as a microwave, the state AOA is predicted instead of NQ and thus a wrong question is generated by \system. Similarly, sometimes an object is not detected by the caption generation model due to poor lighting, partial occlusion, and uncommon texture. In~\Cref{fig:fail_NF_AM}, there is a grey sink made of stone and for the given instruction the state AM would be appropriate. But as the sink is not detected at all, the state NF is wrongly predicted. Sometimes localization errors lead to incorrect state prediction. In \Cref{fig:fail_NQ_AOA}, the two vertically stacked pillows are detected within the same bounding box. Although this is a case of ambiguity, it is predicted as NQ. Additionally, a few incorrect predictions occur due to parsing failure, e.g., for the instruction ``bring me a roll of toilet paper'', the token `toilet' is incorrectly predicted as the \textit{object} class instead of the whole text span, `toilet paper'.

\subsection{User study}
\begin{figure}
    \centering
     \begin{subfigure}[b]{0.7\linewidth}
         \centering
         \includegraphics[width=\textwidth]{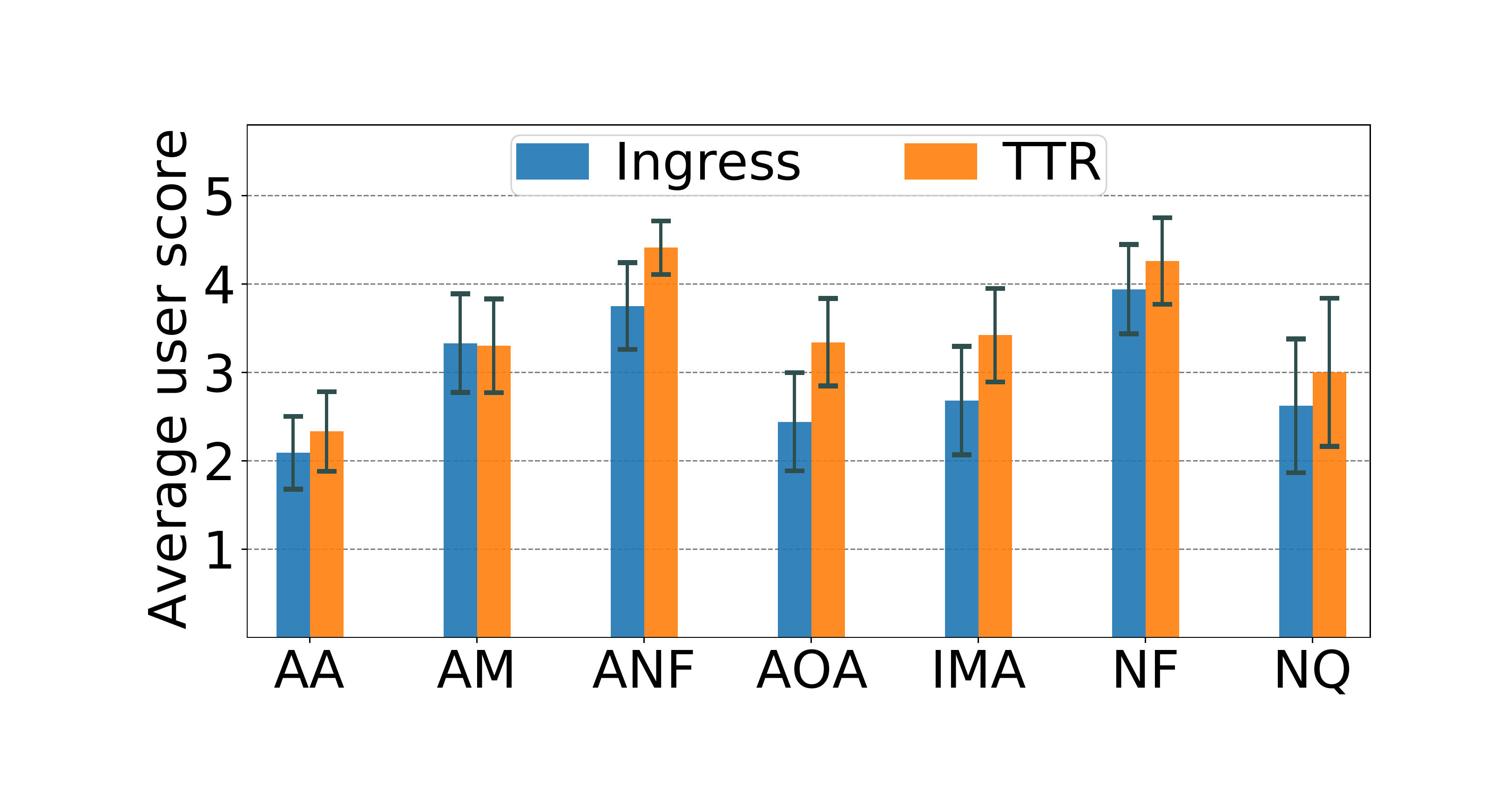}
        \caption{correctness across ambiguity states}
        \label{fig:class_correct}
     \end{subfigure}
     \hfill
     \begin{subfigure}[b]{0.7\linewidth}
         \centering
         \includegraphics[width=\textwidth]{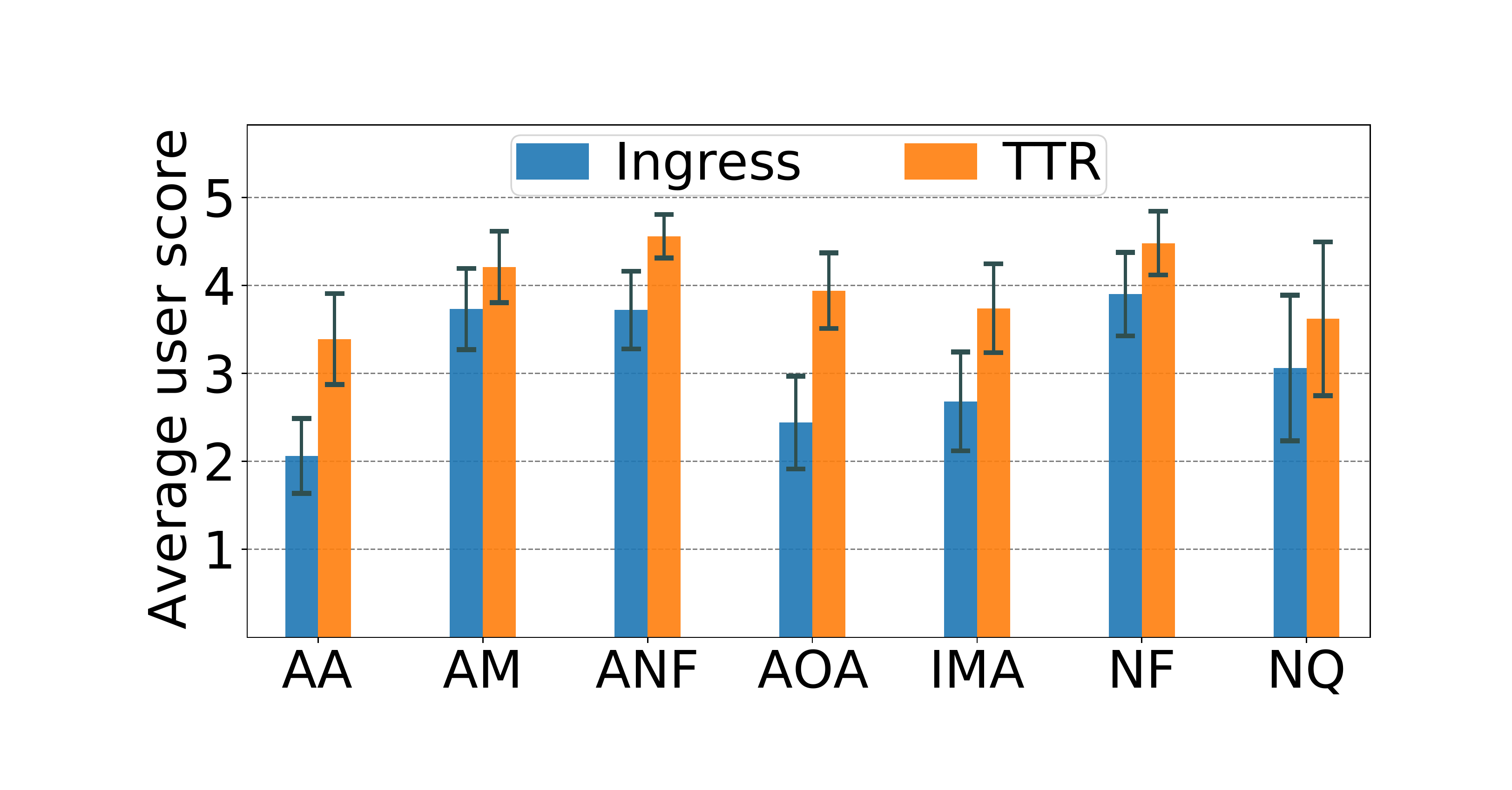}
        \caption{naturalness across ambiguity states}
        \label{fig:class_natural}
     \end{subfigure}
     \hfill
     \begin{subfigure}[b]{0.7\linewidth}
         \centering
         \includegraphics[width=\textwidth]{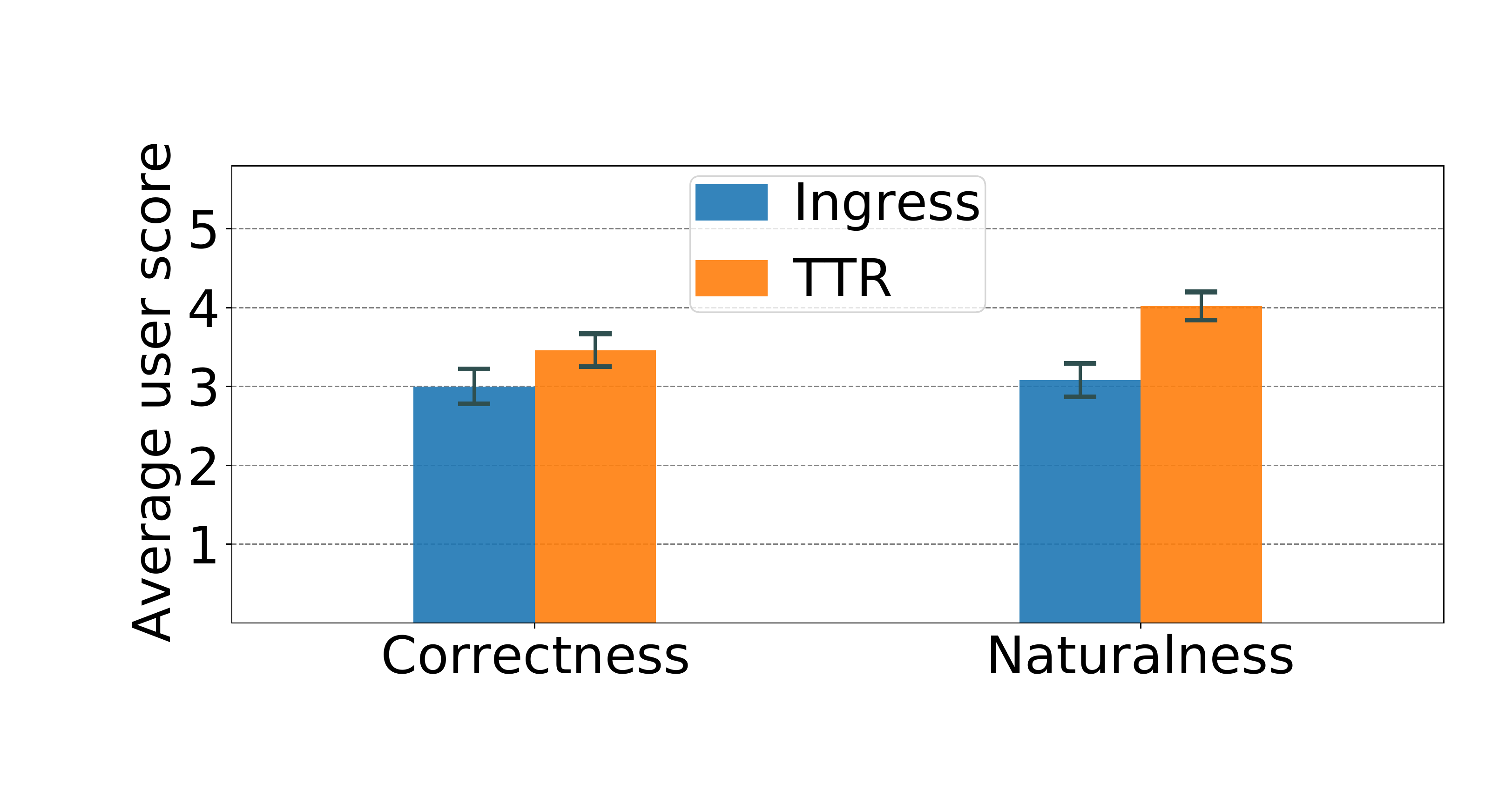}
    \caption{all states combined}
    \label{fig:combined}
     \end{subfigure}
    \caption{From user experience study, average score (5==Best) at 95\% confidence intervals for correctness and naturalness across different ambiguity states and all states combined.}
    \label{fig:user_study}
\end{figure}
We have conducted a small-scale user study, specifically to evaluate the actual questions generated by \system. We compare our natural language question generation with \textit{Ingress}, proposed by Shridar and Hsu~\cite{shridhar18}. To the best of our knowledge, their work is most similar to ours, as they also focus on asking questions for visual disambiguation of objects in a human-robot dialogue scenario and also use the same caption generation model. However, as INGRESS is expected to be used only in a tabletop object manipulation scenario, we perform two changes to the original system to enable the user study. Firstly, as \system produces perspective-free referring expressions to describe an object, we do not use the relational LSTM for question generation that requires the robot and the human to face each other for dialogue. By using perspective-free object descriptions for both systems, we also ensure the user remains unbiased while providing the ratings. This exclusion only affects the question text and not the ambiguity detection, as separate bounding boxes of the same object class are considered as multiple instances in our experiments. Secondly, instead of the robot pointing to different objects in case of ambiguity, we only ask a single question mentioning all the instances, similar to \system. We do so by following the question template in~\cite{shridhar18} and joining multiple referring expressions by the word `or', e.g., ``Do you mean the blue pillow or the yellow pillow?". Following the system description in~\cite{shridhar18} we use the same \densecap model for grounding object descriptions, i.e., the S-LSTM in~\cite{shridhar18}, followed by computing the CEL and METEOR score for the k-means clustering~\cite{shridhar18}.

In this study, a participant is shown multiple image-instruction pairs to evaluate. The participant assesses a given scene and then rates the generated questions in response to the instruction. The participants are asked to rate two questions per image, one generated by \system and another by INGRESS. The participant rates both the questions on a semantic differential scale with numeric points 1-5. We have asked the participants to give their ratings about how correct and natural they perceived the questions to be for the scenario. On the first scale, a 5 rating denotes a completely correct question and 1 denotes a completely incorrect question. Whereas for the second scale, a 5 rating denotes a human-like natural question, and 1 denotes a completely unnatural question. Please note that in the case of samples drawn for the NQ state, the participants are shown an acknowledgment from the robot such as, ``Ok" or ``I can do this'' instead of a question, provided that the state is predicted correctly. A total of 17 participants (9 male, 8 female) from our organization have volunteered for the study. The participants are from the age group 23-38, and all of them are fluent English speakers having at least a bachelor's degree. Four of them have been familiar with the broad research area of robotics, but none of them have any expertise in human-robot interaction and dialogue systems. Each participant repeats the rating process for 14 random image-instruction pairs from our dataset. During the study, we did not reveal which question is generated by which system.

Figure~\ref{fig:class_correct} shows the average correctness ratings for the ambiguity states. The questions generated by \system are perceived to be more accurate for the image-instruction pairs that belong to the ANF, AOA, and IMA states. This result is most likely due to the absence of the fine-grained ambiguity states in INGRESS. INGRESS uses generic questions to tackle different scenarios in a similar way, which possibly impacts the correctness perceived by the participants. Also, the questions generated by \system are perceived to be more natural across all the ambiguity states, as shown in Figure~\ref{fig:class_natural}. We show the overall rating comparison in Figure~\ref{fig:combined}, where the average correctness rating for TTR is 3.46 (SD=1.52) and for Ingress is 3.0 (SD=1.63). Also, the average naturalness rating for TTR is 4.02 (SD=1.32) and for Ingress is 3.08 (SD=1.56). A paired two-tailed t-test reveals the results are statistically significant for both perceived correctness and naturalness. For correctness ratings, the \textit{t} value is 4.79, $p<0.00001$. For naturalness ratings, the \textit{t} value is 9.27, $p<0.00001$.
\section{Discussion}
In this section, we provide some discussion points that would be helpful to adapt the proposed work and perform future research.
\begin{itemize}
    \item In this article, we have analyzed the possibilities of ambiguity in the problem setting and introduced the ambiguity states that are generic and comprehensive. We have also defined a question template associated with each of the ambiguity states. Thus, we reduce the crux of the problem to predicting an ambiguous state and populating the corresponding question template with relevant information. However, predicting the ambiguity states solely from the image and the instruction is difficult, and we have proposed a method that performs this prediction with decent accuracy.
    
    \item To adapt the system to a new domain, the CRF models for task type, argument, and semantic class prediction need to be retrained. However, this does not require a large amount of annotated data. For example, the hyperparameters of the proposed semantic similarity function can easily be fine-tuned with a minimal set of examples, as we use only 10\% of the total dataset for the same. While the question templates may also need to be changed for a different task domain (e.g., UAV for surveillance task), the effort to do so would be minimal as the ambiguity states would be more or less the same. 
    
    \item In regards to End-to-end approaches that directly generate a question, given an image and instruction pair, such a system would not have the fine-grained understanding of the object descriptions. As a result, grounding from the dialogue response of the human would require additional models to parse and match the response to the input. For example, suppose the question generated is ``Do you mean the green or the blue cup’’ and the human responds ``the blue cup’’). We can easily handle the response using the same semantic class prediction model. Moreover, due to the auto-regressive nature of the state-of-the-art text generation models ($P(w_t | w_{t-1}, \dots, w_1 )$), it is generally difficult to control the semantics of the text. Even with beam search decoding, the generation of grammatically correct and semantically incorrect sentences is common. Also, the existing end-to-end models that perform only grounding and no question generation do not handle ambiguity and always choose a single, most appropriate candidate for grounding. In our problem setting, we allow the user to choose the candidate instead of the model. However, with a sufficiently large amount of annotated data, compute power, and research effort, an end-to-end model can be developed in the future.
    
\end{itemize}

\section{Conclusions}
In this article, we describe our \textit{Talk-to-Resolve (TTR)} system that helps a robot in resolving visual ambiguity and inconsistent perception, while executing an instruction. By taking natural language instruction as input, our system analyzes the scene perceived by the robot to convey the exact problem, which helps a human user to correctly signal a confirmation or modification of the task. 
To achieve this, we propose a semantic similarity function to find relevant object description(s) of the scene and a semantic class labeling model to compare the object descriptions with the instruction. Thus, we identify the appropriate ambiguity state so that the exact problem faced by the robot can be expressed as a question. Our experiments suggest that we benefit from our proposed approach for ambiguity state identification in comparison to several baseline systems. We further improve the robustness of our proposed method by suppressing redundant object descriptions. In a user study, we also find that human users perceive the questions from our system to be more accurate and natural, in comparison to the state-of-the-art. Thus, \system provides a significant leap forward in achieving an easy-to-use collocated robotic system in any indoor space.



\bibliographystyle{elsarticle-num} 
\bibliography{main}


\end{document}